\documentclass[twoside,twocolumn,10pt]{article}

\usepackage[left=1.5cm, right=1.5cm, top=2.0cm, bottom=2.0cm]{geometry}
\usepackage[super,sort&compress,comma]{natbib}
\usepackage[version=3]{mhchem}
\usepackage{amsmath,amssymb,mathrsfs}
\usepackage{graphicx}
\usepackage{booktabs}
\usepackage{siunitx}
\usepackage{algorithm}
\usepackage{algpseudocode}
\usepackage{float}
\usepackage{placeins}
\usepackage{stfloats}
\usepackage[usenames,dvipsnames]{xcolor}
\usepackage[T1]{fontenc}
\usepackage{lmodern}      % full T1 font coverage, fixes \micro symbol
\usepackage{textcomp}     % additional symbols including proper µ
\usepackage[english]{babel}
\usepackage{hyperref}
\usepackage{setspace}
\usepackage[compact]{titlesec}
\usepackage{array}
\usepackage[format=plain,justification=justified,singlelinecheck=false,
            font={small},labelfont=bf,labelsep=space]{caption}
\usepackage{epstopdf}
\usepackage{ulem}  % for \sout track-change markup

\addto{\captionsenglish}{%
  
}

\title{\textbf{Learning Pore-scale Multiphase Flow from 4D Velocimetry}}

\author{
  Chunyang Wang,$^{a}$
  Linqi Zhu,$^{a}$
  Yuxuan Gu,$^{a}$
  Robert van der Merwe,$^{b}$
  Xin Ju,$^{c}$\\
  Catherine Spurin,$^{c}$
  Samuel Krevor,$^{a}$
  Rex Ying,$^{d}$
  Tobias Pfaff,$^{e}$
  Martin J. Blunt,$^{a}$\\
  Tom Bultreys$^{b}$
  and Gege Wen$^{a}$\\[1em]
  \small$^{a}$~Department of Earth Science and Engineering, Imperial College London.\\
  \small$^{b}$~Department of Geology, Ghent University.\\
  \small$^{c}$~Department of Energy Science and Engineering, Stanford University.\\
  \small$^{d}$~Department of Computer Science, Yale University.\\
  \small$^{e}$~NVIDIA.\\[0.5em]\\
}

\date{}

\begin{document}

\maketitle

\begin{abstract}
Multiphase flow in porous media underpins subsurface energy and environmental
technologies, including geological CO$_2$ storage and underground hydrogen storage,
yet pore-scale dynamics in realistic three-dimensional materials remain difficult to
characterize and predict. Here we introduce a multimodal learning framework that
infers multiphase pore-scale flow directly from time-resolved four-dimensional (4D)
micro-velocimetry measurements. The model couples a graph network simulator for
Lagrangian tracer-particle motion with a 3D U-Net for voxelized interface evolution.
The imaged pore geometry serves as a boundary constraint to the flow velocity and the
multiphase interface predictions, which are coupled and updated iteratively at each
time step. Trained autoregressively on experimental sequences in capillary-dominated
conditions ($Ca\approx10^{-6}$), the learned surrogate captures transient, nonlocal
flow perturbations and abrupt interface rearrangements (Haines jumps) over rollouts
spanning seconds of physical time, while reducing hour-to-day--scale direct numerical
simulations to seconds of inference. By providing rapid, experimentally informed
predictions, the framework opens a route to ``digital experiments'' to replicate
pore-scale physics observed in multiphase flow experiments, offering an efficient tool
for exploring injection conditions and pore-geometry effects relevant to subsurface
carbon and hydrogen storage.
\end{abstract}

%%%=============================================================
%%% MAIN TEXT
%%%=============================================================

\section*{Introduction}

Multiphase transport in porous media is fundamental to a broad class of natural and
engineered systems. It is central to the efficiency and security of geological carbon
sequestration\cite{Siegelman_Kim_Long_2021, Chen_Lu_Liao_Chen_Li_Wang_Lv_Cui_Lan_Wang_et_al._2025, doi:10.1073/pnas.1918837117, epic37530}
and underground hydrogen
storage,\cite{Oh_Tumanov_Ban_Li_Richter_Hudson_Brown_Iles_Wallacher_Jorgensen_et_al._2024, Kim_Kim_Kim_Kwon_Jin_Ha_Shim_Park_Jamal_Kim_et_al_2024}
and plays an important role in related technologies such as fuel-cell water
management\cite{Wang_Zhang_Favero_Higgins_Luo_Stephens_Titirici_2024} and nuclear
waste containment.\cite{Li_Dong_Wang_Ma_Tan_Jensen_Deibert_Butler_Cure_Shi_etal._2017}
Across these applications, a substantial portion of transport occurs in
capillary-dominated regimes (capillary number $Ca \leq 10^{-6}$), where pore-scale
interface dynamics, geometry, and nonlocal flow perturbations exert a first-order
control on macroscopic behavior. Accurate prediction of such flows is therefore
essential for optimizing system performance while mitigating environmental and societal
risks.\cite{Mouli-Castillo_Wilkinson_Mignard_McDermott_Haszeldine_Shipton_2019, C7EE02342A}
Fluid flow in porous media is governed by the Navier-Stokes equations,\cite{tryggvason2011direct}
and the capillary forces at fluid--fluid interfaces are described by the Young-Laplace
equation.\cite{lake1989enhanced} Although the underlying equations for multiphase flow
through porous media are well established, the practical numerical modeling of
multiphase behavior remains a widely recognized challenge due to the complexity of
pore-scale interactions and the absence of accurate closure relations across
scales.\cite{Armstrong_McClure_Berrill_Rücker_Schlüter_Berg_2016, Zhang_YH_2021}

%%% FIGURE 1 %%%
\begin{figure*}[!htbp]
  \centering
  \includegraphics[width=\textwidth]{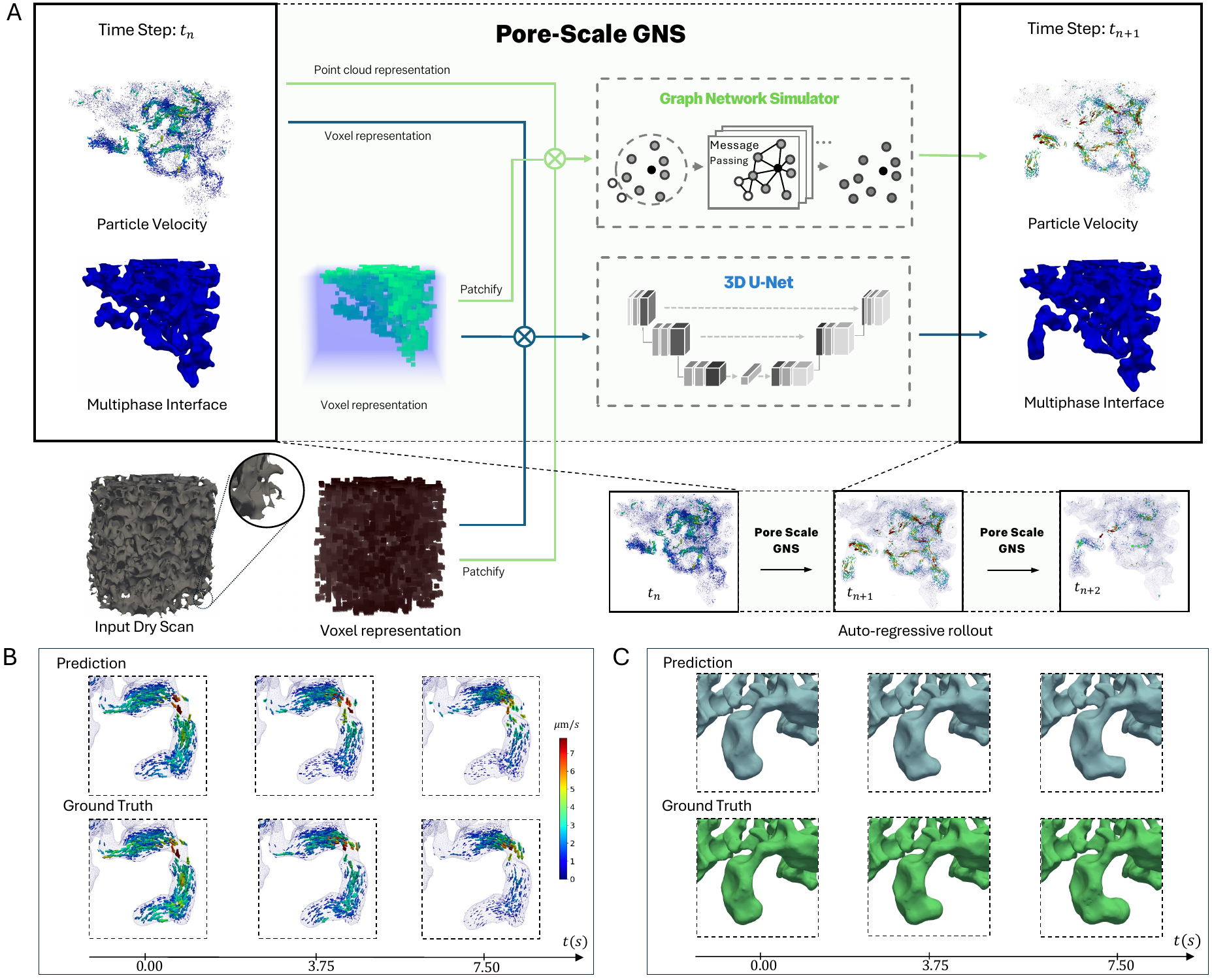}
  \caption{
    \textbf{Overview of the Pore Scale GNS framework for learning multiphase fluid
    dynamics in porous media from 4D velocimetry experimental data.}
    \textbf{(A)} At each time step, the model takes as input the particle velocity
    field and the multiphase interface, both obtained from high-resolution 4D
    micro-velocimetry experiments, alongside the static dry image of the pore geometry.
    These inputs are processed through a multimodal architecture combining a graph
    neural network and a 3D U-Net. The model is trained to autoregressively predict
    future states, enabling long-horizon rollout of pore-scale multiphase flow.
    \textbf{(B)} Representative velocity field predictions over time compared to ground
    truth, showing the accurate reconstruction of the spatiotemporal flow structures.
    \textbf{(C)} Corresponding predicted versus actual multiphase interface evolution,
    demonstrating the model's ability to recover dynamic interfacial morphology.
  }
  \label{fig:main}
\end{figure*}

A key difficulty lies in the spatially and temporally multiscale nature of the
multiphase problem, governed by tightly coupled inertial, viscous, capillary, and
interfacial forces.\cite{Singh_Jung_Brinkmann_Seemann_2019, PhysRevLett.111.064501, datta2014fluid, carrillo2020multiphase}
Both transient instabilities and persistent sub-pore flow mechanisms --- including
Haines jumps, wetting layer flow, contact line dynamics, and thin-film breakup ---
significantly influence larger-scale flow
behavior.\cite{Ferrari_Jimenez-Martinez_Borgne_Méheust_Lunati_2015, doi:10.1073/pnas.1901619116}
However, explicitly representing the effect of these sub-pore phenomena in continuum
models proves very challenging. Long-range
interactions\cite{doi:10.1073/pnas.1221373110, PhysRevE.88.043010, Mansouri-Boroujeni_Soulaine_Azaroual_Roman_2023}
of the flow velocity and multiphase fluid interface prevent a clean decomposition of
the dynamics into independent spatial and temporal scales. Multiphase pore-scale flow
is commonly investigated using pore-network models, which efficiently explore
topology-controlled displacement but rely on quasi-static invasion rules and idealized
throat geometries,\cite{Blunt_2001} and mesoscopic lattice Boltzmann methods, which
directly resolve interfacial hydrodynamics but are constrained by diffuse-interface
assumptions and limited density ratios.\cite{Pan_Hilpert_Miller_2004} Both classes of
solver face a fundamental nano-to-centimeter scale gap: faithfully representing
sub-pore effects requires boundary conditions at nanometer resolution over
millimeter--centimeter domains---a constraint that remains computationally
intractable. Experimental observations, by contrast, intrinsically integrate these
sub-pore effects at the observed (micrometer) scale, providing physically consistent
data that encode the cumulative influence of fine-scale processes without requiring
their explicit resolution.\cite{https://doi.org/10.1029/2023WR034720}

Beyond numerical modeling, micro-CT imaging has emerged as a critical approach for
understanding multiphase pore-scale phenomena by enabling direct \textit{in situ}
observation of flow dynamics in realistic porous
materials.\cite{Wildenschild_Sheppard_2013} Modern synchrotron-based facilities now
achieve micrometer-scale spatial resolution (\SI{}{\micro\meter}) and sub-second
temporal resolution (\SI{}{\milli\second}),\cite{Gorenkov_Nikitin_Fokin_Duchkov_2024}
leading to key experimental insights such as intermittent gas and liquid transport
pathways and previously unresolved mechanisms of phase
redistribution.\cite{spurin2019intermittent} When combined with tracer particles,
synchrotron-based 4D micro-velocimetry further enables simultaneous, time-resolved
measurements of pore-scale velocity fields and fluid--fluid interface evolution,
revealing long-range flow responses associated with Haines jumps and other transient
interfacial events.\cite{tom4d} Additionally, these measurements also offer the
spatiotemporal resolution ideal for training machine learning models to directly learn
pore-scale multiphase dynamics.

%%% FIGURE 2 %%%
\begin{figure*}[!htbp]
  \centering
  \includegraphics[width=\textwidth]{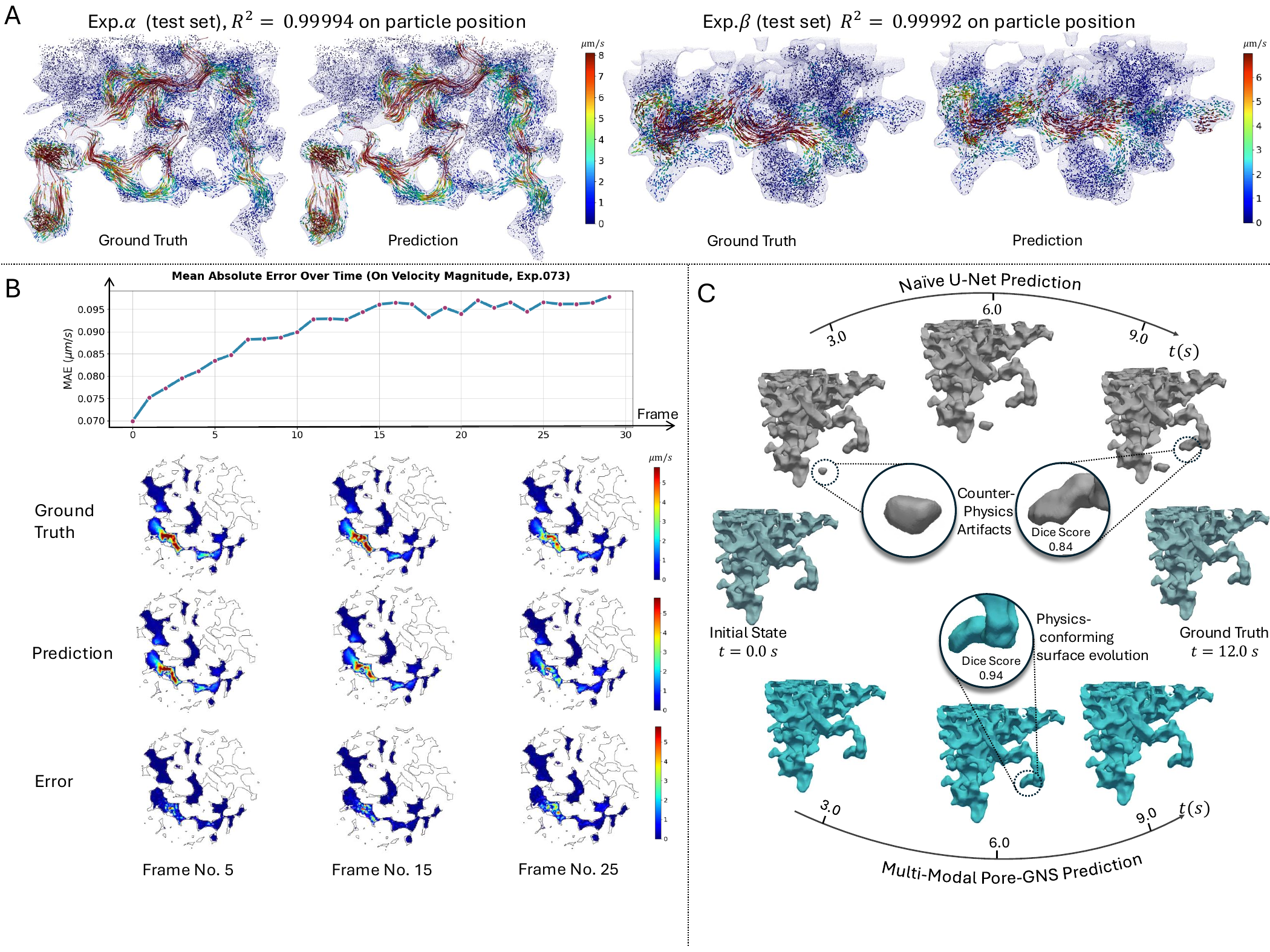}
  \caption{
    \textbf{Multimodal prediction of particle-scale flow and fluid interface dynamics.}
    \textbf{(A)} Comparison of predicted and ground truth particle trajectories for two
    different meniscus configurations. Left: Experiment $\alpha$, Right: Experiment
    $\beta$. $R^{2}$ measured on velocity tracer position prediction for 30 time steps.
    \textbf{(B)} Temporal evolution of velocity error across 30 frames (top), along
    with frame-wise comparison of predicted and ground truth flow fields in a
    representative slice for Exp.$\alpha$. Error maps highlight that most discrepancies
    arise near high-gradient regions.
    \textbf{(C)} Comparison between na\"ive architecture inference (top path) and
    multimodal prediction (bottom path) from the same initial interface. The na\"ive
    method, relying solely on surface history, introduces nonphysical artifacts (upper
    center). In contrast, our approach incorporates particle velocity, a complementary
    modality, to guide interface prediction, effectively avoiding artifacts and
    producing a physically-consistent geometric surface. Regional Dice Score is
    reported for the Haines jump area.
  }
  \label{fig:result}
\end{figure*}

Machine learning (ML) has demonstrated strong capabilities for learning complex
phenomena directly from data, with broad successes in natural language
processing,\cite{brown2020language} computer vision,\cite{he2016deep} and
computational biology.\cite{jumper2021highly} In fluid mechanics, data-driven
surrogates---including convolution-based models such as
FluidNet\cite{Yaqoob_Ansari_Ishaq_Ashraf_Pavuluri_Rabbani_Rabbani_Seers_2025} and
particle-based Graph Network Simulators (GNS)\cite{10.5555/3524938.3525722}---have
achieved orders-of-magnitude speedups over conventional solvers. However, applying
these architectures to porous media poses two key challenges: most existing ML
architectures are designed for regular domains or simple
geometries,\cite{wen2023real, bodnar2025foundation, li2023fourier} limiting their
applicability to irregular pore topologies; and reliable training data from pore-scale
numerical simulations remain scarce because current solvers often simplify or omit
critical interfacial and topological
complexities.\cite{Yang2024review} High-resolution experimental 4D micro-velocimetry
datasets offer a route to overcome both limitations, providing rich, physically
consistent observations from which models can learn directly.

Here we introduce the Pore-Scale Graph Network Simulator (Pore-Scale GNS)
(Fig.~\ref{fig:main}), the first ML architecture to directly learn multiphase flow
physics from experimental 4D velocimetry data in porous
media.\cite{tom4d} The framework couples a graph network simulator for Lagrangian
tracer-particle dynamics with a 3D U-Net for voxelized interface evolution in a
two-way multimodal architecture, using the imaged pore geometry as a boundary
constraint and as a shared spatial conditioning input for both velocity and interface
predictions. The GNS represents tracer particles as nodes in a dynamically
reconstructed graph, with edges connecting neighbors within a physically motivated
radius; iterative message passing then propagates local flow information across the
network, naturally encoding the long-range velocity correlations that are characteristic
of capillary-dominated displacement---a feature that isotropic grid-based methods
cannot readily replicate. The 3D U-Net complements this by operating on the voxel
grid: its encoder--decoder structure with skip connections is well-suited to capturing
the multiscale spatial structure of evolving binary interfaces, and its channel design
allows the interpolated GNS velocity field to be ingested directly as a physical
conditioning signal, coupling the Lagrangian particle description to the Eulerian
interface representation. Trained autoregressively on capillary-dominated drainage
sequences, the model generalizes effectively to unseen flow configurations within the
same porous medium and, in a zero-shot (no fine-tuning on the target medium) setting,
to a substantially different pore structure. These results demonstrate that learning
directly from 4D experimental observations can yield physically grounded surrogate
models for multiphase transport, opening a path toward rapid ``digital experiments''
for subsurface energy applications.

\section*{Results and Discussion}

We design the multimodal ML framework, Pore-Scale GNS (Fig.~\ref{fig:main}A), to learn
multiphase flow dynamics directly from experimental data. Trained on time-resolved 4D
micro-velocimetry sequences and evaluated on both temporal extrapolation and
structurally distinct test samples, the model recovers particle-scale velocities
(Fig.~\ref{fig:main}B) and fluid--fluid interface evolution
(Fig.~\ref{fig:main}C). The experimental data captures slow,
capillary-dominated drainage in a sintered glass porous medium (capillary number
$Ca \approx 10^{-6}$,\cite{tom4d}). It was acquired at the TOMCAT beamline using
high-speed synchrotron micro-CT, providing isotropic voxels of \SI{2.75}{\micro\meter}
and a field of view of 5.5 $\times$ 5.5~mm$^2$. Time-resolved 3D tomograms were
captured at 0.25 to \SI{0.5}{\second} per frame in bursts of up to 200~scans,
simultaneously resolving fluid--fluid interfaces and tracer particle velocities during
drainage events.

%%% TABLE 1 %%%
\begin{table*}[!htb]
  \centering
  \caption{
    Evaluation metrics for fluid topology and velocity prediction across different
    datasets and rollout settings. Volume and surface area errors are reported as
    relative percentage errors, while the Dice score quantifies voxel-wise agreement.
    The velocity field is interpolated based on the predicted tracer velocity. The
    lower MAE in Exp.$\beta$ compared to Exp.$\alpha$ is attributed to its lower mean
    velocity magnitude. The trajectory $R^2$ measures the coefficient of determination
    between predicted and ground truth particle positions. All metrics for rollout test
    sets are averaged over time.
  }
  \label{tab:surf_metrics}
  \small
  \begin{tabular*}{\textwidth}{@{\extracolsep{\fill}}lrrrrr}
    \hline
    \textbf{Experiment Setup} & \textbf{Vol. Err. (\%)} & \textbf{Surf. Err. (\%)} & \textbf{Dice} & \textbf{Vel. MAE (\SI{}{\micro\meter\per\second})} & \textbf{Traj. $R^{2}$} \\
    \hline
    Exp.$\alpha$ - Train, One-Step (30 Fr.) & 0.174 & 0.220 & 0.990 & 0.707 & 0.9999 \\
    Exp.$\alpha$ - Test, One-Step (30 Fr.)  & 0.187 & 0.348 & 0.989 & 0.738 & 0.9999 \\
    Exp.$\alpha$ - Train, Rollout (30 Fr.)  & 0.198 & 0.221 & 0.985 & 0.919 & 0.9999 \\
    Exp.$\alpha$ - Test, Rollout (30 Fr.)   & 0.351 & 0.613 & 0.986 & 1.000 & 0.9999 \\
    \hline
    Exp.$\beta$ - Test, One-Step (30 Fr.) & 0.683 & 1.221 & 0.991 & 0.098 & 0.9999 \\
    Exp.$\beta$ - Test, Rollout (30 Fr.)  & 0.735 & 1.324 & 0.990 & 0.119 & 0.9999 \\
    \hline
  \end{tabular*}
\end{table*}

Two distinct datasets from the same sample are used: Exp.$\alpha$ and $\beta$.
Exp.$\beta$ images the lower subvolume of the sintered-glass sample, whereas
Exp.$\alpha$ covers the upper subvolume; since the menisci are located in different
pore spaces across the two subvolumes, this constitutes a boundary-condition
generalization test: the pore geometry is statistically identical between subvolumes,
but the meniscus configuration, initial saturation, and mean flow rate differ. In
addition, Exp.$\beta$ features lower mean flow velocities than Exp.$\alpha$.
Exp.$\alpha$ (initial saturation for nonwetting phase $S^{\alpha}_{nw} = 0.3518$) is
used for training: the first 150 frames are provided as input, and the model is rolled
forward autoregressively to predict the next 30 frames ($\sim$7.5~s) as an
extrapolation test set. Exp.$\beta$, with an initial saturation for the nonwetting
phase $S^{\beta}_{nw} = 0.3576$, occurs prior to Exp.$\alpha$ in physical time. It
is originally acquired at a resolution of 0.5 seconds per frame. Through temporal
interpolation (refer to Appendix~\ref{sec:sdf_interp}), it is resampled to 0.25
seconds per frame and utilized as a test case.

\subsection*{Velocity Prediction Accuracy and Stability over Time}

Velocity prediction results reported are evaluated on Exp.~$\alpha$, with the model
rolled out autoregressively beyond the training window
(Table~\ref{tab:surf_metrics}). The trained Pore-Scale GNS shows excellent accuracy
for velocity prediction during both training and the extrapolated inference period. As
shown in Fig.~\ref{fig:result}, we evaluate the accuracy of the particle velocity
prediction using two methods: streamlines and the interpolated velocity field. The
predicted streamline plot in Fig.~\ref{fig:result}A has highlighted trajectories of
particle tracers flowing into the Haines jump region. The streamline tracer particle
location prediction achieved an $R^2$ accuracy of 0.9999 over 30~extrapolation time
steps.

Fig.~\ref{fig:result}B shows that the inference mean absolute error (MAE) of the
interpolated velocity field increases gradually over time, but stabilizes below
\SI{1.1}{\micro\meter\per\second}. Importantly, no significant particle trespassing
across the fluid--fluid interface is observed during the autoregressive rollout. This
is direct evidence of an effective multimodal design: the interface geometry predicted
by the U-Net is explicitly coupled into the GNS velocity prediction at each step,
ensuring that particle motion respects the evolving phase boundary
(see Fig.~\ref{fig:geo_cmp}A). This integration of physical knowledge into the
architecture prevents spurious cross-phase transport and contributes to the observed
stability over long rollout horizons.

\subsection*{Interface Reconstruction and Topological Accuracy}

We evaluate the quality of the prediction of the dynamic fluid--fluid interface for
Exp.$\alpha$ using three complementary metrics: volume relative error, surface area
relative error, and the Dice similarity coefficient\cite{muller2022guideline}
(Table~\ref{tab:surf_metrics}). Across both training and test datasets, and for both
single-step and long-horizon (30-frame) rollouts, the model demonstrates consistently
low volume and surface errors, with Dice scores exceeding 0.98. In the 30-frame
rollout, the volume error remains below 0.4~\%, and the surface area error is
similarly low, thereby demonstrating the model's capacity to reconstruct physically
meaningful interface morphology over time.

A key factor in this performance is the multimodal design of the framework, in which
particle velocity predictions from the GNS are incorporated into the U-Net interface
reconstruction. Without this conditioning, counter-physical artifacts such as spurious
phase emergence in void regions appear, and Dice accuracy deteriorates in regions of
rapid interfacial motion. By contrast, including velocity information improves Dice
scores in critical regions where Haines jumps occur by 10.6\%
(Fig.~\ref{fig:result}C). To achieve this, particle velocities are sampled onto a
voxel grid and downsampled to the U-Net input size using the max-pooling scheme shown
in Fig.~\ref{fig:patchsize}B, which preserves over 99\% of the information. This
approach avoids the severe information loss that typically arises from repeated slicing
in three dimensions, ensuring that velocity--interface coupling is maintained at high
fidelity.

\subsection*{Generalization across Experiments}

Generalization is assessed at three distinct levels: (a)~temporal extrapolation beyond
the training window within Exp.$\alpha$; (b)~boundary-condition generalization to a
different flow state within the same sintered-glass pore geometry (Exp.$\beta$); and
(c)~zero-shot cross-rock transfer to a structurally distinct medium (Ketton limestone;
\textit{Generalization across Rock Types}). For level (b), we evaluated the trained
model on an unseen experiment (Exp.$\beta$) that differs from the training case
(Exp.$\alpha$) in meniscus configuration, mean flow rate, and initial saturation,
while sharing statistically identical sintered-glass pore geometry. Exp.$\alpha$
features more tortuous pathways with higher average flow velocities, while Exp.$\beta$
contains a smaller field-of-view and lower average velocities. As shown in
Fig.~\ref{fig:result}A, the model reconstructs velocity streamlines for the unseen
experiment with high fidelity, achieving an $R^{2}$ score of 0.9992 on Exp.$\beta$,
measured by predicted particle position---comparable to in-distribution
performance---though modest increases in interface volume and surface area errors are
observed (Table~\ref{tab:surf_metrics}).

The mean MAE of the velocity field is \SI{1.000}{\micro\meter\per\second} for
Exp.$\alpha$ and \SI{0.119}{\micro\meter\per\second} for Exp.$\beta$. These absolute
values reflect the substantially different characteristic flow speeds of the two
experiments: Exp.$\beta$ operates at considerably lower velocities, so an
order-of-magnitude lower MAE is consistent with comparable relative predictive
accuracy. These results suggest that the framework captures generalizable features of
multiphase geometry and transport within the same porous medium.

We further validate this generalization under a domain shift by performing strict
zero-shot inference on a different rock type: the model trained only on the
sintered-glass dataset is applied to an independent drainage experiment in Ketton
limestone without fine-tuning. The porous media differ substantially in their
void-space structure (mean pore size 122~\SI{}{\micro\meter} and image-based porosity
27\% for the sintered glass filter versus 52~\SI{}{\micro\meter} and 14\% for Ketton).
The experiments use the same wetting/nonwetting pair (KI-brine and silicone oil), and
both samples are water-wet relative to oil, but the solid surface chemistry differs
(glass versus calcite).

\begin{figure}[!tb]
  \centering
  \includegraphics[width=\columnwidth]{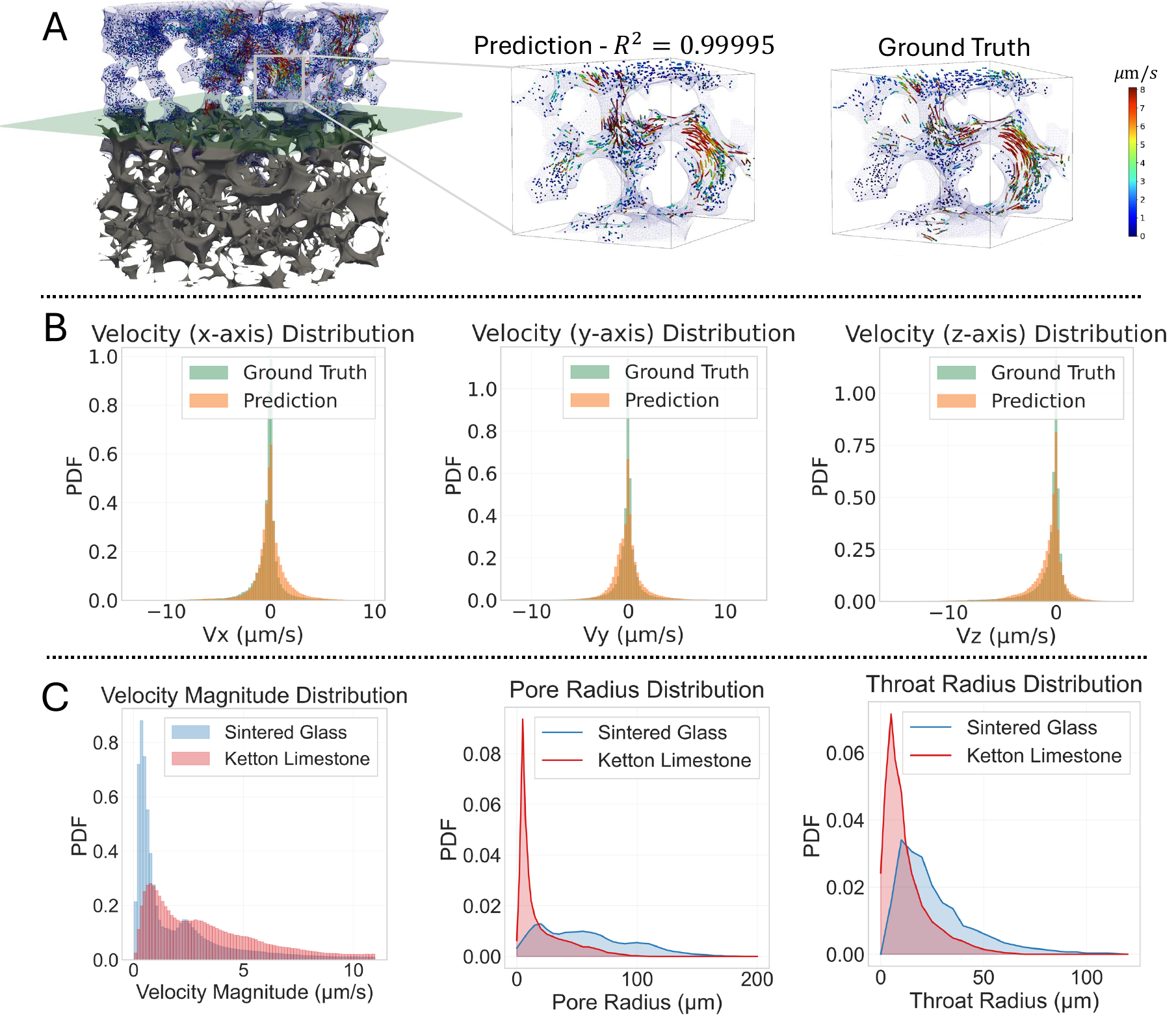}
  \caption{
    \textbf{Zero-shot generalization to a new rock type (Ketton).}
    \textbf{(A)} Predicted versus ground-truth particle trajectories and velocity
    visualization (including a zoomed-in view of the pore space) for an independent
    drainage experiment in Ketton limestone, evaluated strictly zero-shot (no
    fine-tuning). The reported $R^{2}$ is computed from agreement in particle positions
    along trajectories.
    \textbf{(B)} Velocity distribution comparing model predictions with ground truth
    for Ketton, used to quantify performance under the cross-rock shift.
    \textbf{(C)} Domain-shift characterization: distributions of key pore-scale
    descriptors for the training medium (Sintered glass) versus the new medium
    (Ketton), including pore radius, throat radius, and velocity-magnitude statistics.
  }
  \label{fig:ketton}
\end{figure}

With the cross-rock domain shift, predictive performance degrades relative to the
in-distribution sintered-glass cases, though the model retains partial predictive
capability as detailed below. Fig.~\ref{fig:ketton}A shows physically plausible
particle trajectories and velocity structures in the Ketton limestone sample, including
within complex pore regions highlighted in the zoomed-in view. At the domain level,
predicted and ground-truth velocity statistics show reasonable agreement, though with
larger discrepancies than observed for the in-distribution sintered-glass cases.
Fig.~\ref{fig:ketton}C compares velocity-magnitude distributions; to quantify
prediction accuracy we adopt the normalised root-mean-square error relative to the
99th-percentile characteristic velocity ($\mathrm{NRMSE_{p99}}$; see
Appendix~\ref{sec:nrmse_metric} for definition and rationale), which is robust to the
large fraction of near-stagnant particles present in capillary-dominated drainage. The
model achieves $\mathrm{NRMSE_{p99}} = 16.4\%$ on the Ketton dataset, indicating
partial but non-trivial statistical agreement with the experimental velocity field
under the cross-rock domain shift. Compared to the corresponding in-distribution case
(Exp.$\alpha$), where $\mathrm{NRMSE_{p99}} = 12.5\%$, the error increases to 16.4\%
for Ketton, consistent with the expected degradation under cross-rock transfer.
Alongside this partial agreement, localized degradation occurs in specific pore
regions, as highlighted in the failure-mode analysis
(Fig.~\ref{fig:failure}, Appendix~\ref{sec:failure_modes}), where reduced trajectory
alignment and velocity accuracy reflect the limits of strict zero-shot transfer under
substantial pore-structure differences. Fig.~\ref{fig:ketton}B further contextualizes
this behavior by quantifying domain shifts in pore-scale descriptors between sintered
glass and Ketton. Ketton inference uses a model trained on an expanded sintered-glass
dataset with two additional sequences
(Fig.~\ref{fig:extra_exp}, Appendix~\ref{sec:supp_training}), improving in-domain
robustness while preserving strict zero-shot transfer.

%%% FIGURE 3 %%%

\subsection*{Computational Efficiency}

Computational cost for pore-scale numerical simulations depends strongly on domain
size, capillary number, initialization scheme, and boundary conditions. For samples of
comparable size and flow regime, the effort typically ranges from $O(10^5)$ to
$O(10^7)$ time steps, with stringent convergence criteria exerting a major influence
on stability and accuracy. Even with substantial computational resources (e.g., 80
parallel NVIDIA V100 GPUs), numerical simulations for cases comparable to ours require
between 2 and 20 hours of wall-clock time, depending on capillary number and
initialization quality.\cite{com_eff,7dc26e4174af481fb979f0e74a1ba654} Critically,
this cost grows most severely at low capillary number, where strong interfacial forces
demand much tighter convergence and additional iterations per timestep.

In contrast, our framework requires only $\sim$200 autoregressive steps to reproduce
an experiment of \SI{50}{\second}; inference runs at a constant rate of
$\sim$0.027~s per step on a single NVIDIA A100 (80~GB), regardless of capillary
number or boundary-condition complexity. For comparable capillary-dominated
conditions, direct numerical simulations require $\mathcal{O}(10^{4}\!-\!10^{5})$
seconds of wall-clock time, whereas our framework requires $\sim 5$ seconds,
corresponding to a $10^{3}\!-\!10^{4}$-fold speedup. For training on the same GPU,
using the first 150 frames of Experiment~$\alpha$, the wall-clock times were 4~h~8~min
for the Graph Network Simulator (GNS) and 48~min for the 3D U-Net. This efficiency
gain unlocks pore-scale predictions that were previously computationally intractable.

%%% FIGURE 4 %%%
\begin{figure}[!tb]
  \centering
  \includegraphics[width=\columnwidth]{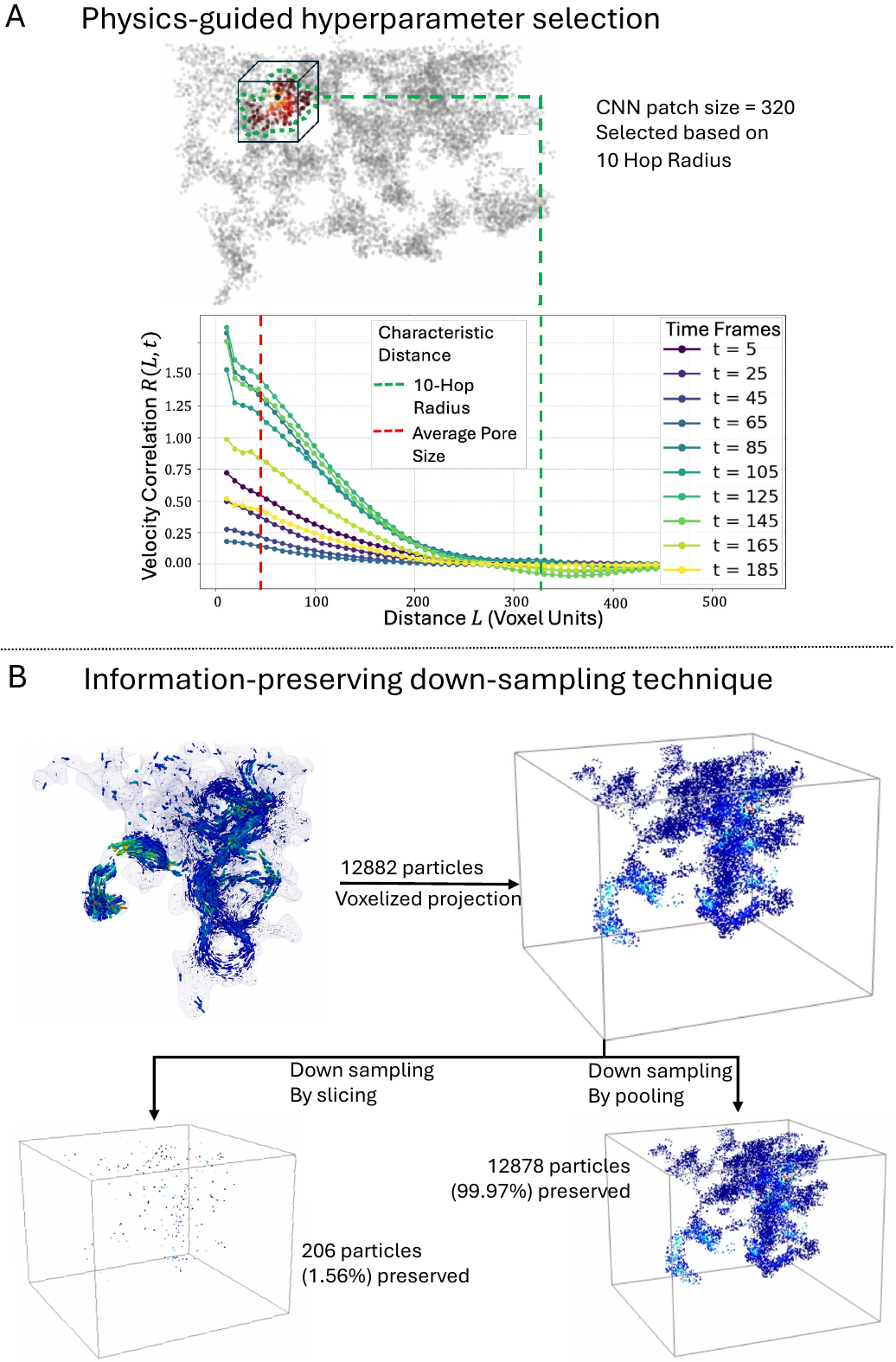}
  \caption{
    \textbf{Multimodal information exchange between geometry and velocity.}
    \textbf{(A)} Information from the predicted interface $\mathcal{S}$ is passed from
    the U-Net surface model to the GNS velocity model by extracting local voxel patches
    centered on particles. The patch size is selected based on experimental velocity
    correlation data, shown in the plot. The GNS model uses a 10-hop graph structure,
    with each hop attending to 32 neighbors, resulting in an effective receptive field
    of 320 voxels, which corresponds to the spatial extent over which velocity
    correlations remain significant.
    \textbf{(B)} Velocity data from particles are transferred to the U-Net by
    projecting the voxelized velocity onto a 3D grid. Pooling-based downsampling
    (downsample factor 4 here for demonstration) is applied to reduce memory cost while
    enabling efficient integration of flow information. Na\"ive downsampling via slicing
    leads to severe information loss due to the sparsity and three-dimensional nature
    of the data (retaining only 1.56\% of particles). In contrast, pooling-based
    downsampling preserves local velocity extremes and retains 99.97\% of particles.
  }
  \label{fig:patchsize}
\end{figure}

\subsection*{Physics-Guided Model Architecture Design}

We found that the architecture design must reflect the tightly coupled relationship of
the fluid--fluid interface geometry and the particle velocity. Ablation studies (see
\textit{Materials and Methods}) show that simpler methods, such as a GNS without
multimodal information, fail to accurately predict the flow; incorporating particle
velocity information enhances the accuracy of interface predictions by 10\% measured
in Dice Score within the Haines jump region (see Fig.~\ref{fig:result}C for a
zoomed-in view).

We also used the particle flow physics to inform the hyperparameter selection, as
shown in Fig.~\ref{fig:patchsize}A. In the GNS, we chose 10 message passing hops and
the local context CNN patch size to be 32 voxels, using the velocity correlation
analysis result, which is $\sim$320 voxels. This design allows the Pore-Scale GNS to
be informed by the long-range flow dynamics to predict the velocity at the next time
step.

\subsection*{Broader Implications}

The predictive power demonstrated in this work can be used to advance the
understanding of mass transfer processes governed by chemical transport, including
fluid--fluid dissolution and mineral dissolution in subsurface environments.
Concentration fields in such systems are inherently heterogeneous: values near
fluid--fluid and fluid--solid interfaces approach solubility limits, whereas bulk
regions remain under-saturated due to incomplete mixing and diffusion
constraints.\cite{Agaoglu_Copty_Scheytt_Hinkelmann_2015} In these regimes, both
local velocity fields and interfacial geometry are key in controlling transfer
efficiency and long-term system evolution.\cite{Agaoglu_Copty_Scheytt_Hinkelmann_2015,YAO2021116017}
Surface morphology, particularly curvature and topology, governs capillary forces,
flow stability, and interphase transfer rates, yet bulk measures such as porosity or
permeability fail to capture these transport-relevant
features.\cite{Armstrong_McClure_Robins_Liu_Arns_Schlüter_Berg_2019} By combining
particle-resolved velocity with voxel-based interface predictions in a unified
framework, our approach provides a physically grounded tool to evaluate phenomena such
as capillary trapping, dissolution efficiency, and interfacial stability.

Beyond the specific systems studied here, the velocity--interface coupling framework is
applicable to a broader class of subsurface and engineered porous systems where
capillary and viscous forces compete, including enhanced oil recovery, reactive
transport during mineral dissolution or precipitation, and contaminant migration in
remediation systems.

\section*{Contribution}

This study shows that learning directly from time-resolved, particle-resolved 4D
micro-velocimetry provides a distinct route to modeling multiphase flow in porous
media. Unlike prior ML efforts that largely relied on static 3D micro-CT to infer
scalar properties such as permeability\cite{Zhao_Han_Guo_Chen_2025} or on 2D
time-series to approximate limited dynamics,\cite{Lee2023SciRep} the present approach
leverages the full information content of 4D datasets where velocities and interfaces
are observed together, yielding more time-resolved and physically grounded predictions.
Generalisation experiments demonstrate that the framework transfers effectively to
distinct boundary conditions and flow states within the same sintered-glass pore
geometry (Exp.$\beta$; boundary-condition generalization), and retains attenuated
predictive capability under strict zero-shot transfer to a structurally distinct rock
type (Ketton limestone; structural generalization), while highlighting the expected
degradation that accompanies pronounced domain shifts.

More broadly, these results position data-driven models trained on rich experimental
measurements as a complementary method to first-principles simulation, enabling rapid
exploration of injection scenarios, pore-geometry variations, and flow regimes that
remain computationally prohibitive for conventional solvers. The limited diversity of
the current training corpus reflects a field-wide constraint: synchrotron-based 4D
micro-velocimetry acquisitions demand substantial dedicated beamtime and specialised
sample preparation, making comprehensive multi-rock, multi-regime datasets difficult
to assemble at this stage of the field's development. The present work, by
demonstrating effective learning from a small number of carefully selected sequences,
is intended in part to substantiate the scientific return on such experimental
investment. Key next steps are to validate generality beyond the present dataset by
expanding to 4D micro-velocimetry acquired across diverse porous materials, wettability
states, and flow conditions, to mitigate long-horizon degradation from autoregressive
error accumulation (Fig.~\ref{fig:failure}, Appendix~\ref{sec:failure_modes}) through
larger, more diverse training data and improved temporal coupling or hybrid
physics-informed constraints, and to connect pore-scale predictions to macroscopic
descriptors such as relative permeability and capillary pressure curves for direct
integration into field-scale subsurface flow models.

\section*{Experimental (Materials \& Methods)}

\subsection*{Dataset and Data Processing}

The dataset used in this work was obtained through the 4D micro-velocimetry
experimental procedure of Bultreys et al.,\cite{tom4d} which provides direct,
time-resolved measurements of pore-scale flow and interface dynamics. This dataset
captures complex pore-scale phenomena, including Haines jumps, long-range velocity
perturbations, and highly tortuous flow paths, providing a rich experimental foundation
for data-driven modeling of multiphase flow.

This experiment tracked the displacement of viscous silicone oil into brine-saturated
sintered glass filters (mean pore size $122~\mu$m, porosity 27\%) using tracer
particles (\SI{10}{\micro\meter} diameter, silver-coated hollow spheres) at a
controlled injection rate corresponding to a capillary number of approximately
$5.5 \times 10^{-6}$. The imaging was performed at the TOMCAT beamline (Swiss Light
Source) with \SI{2.75}{\micro\meter} voxel resolution and 0.25~s temporal resolution,
enabling full 3D reconstruction of both the multiphase fluid interface and Lagrangian
particle trajectories over a 50-second time window. Particles were detected on each
tomogram in the time series using local grey value peak detection. These detections
were subsequently linked into particle trajectories using a location-prediction-based
nearest-neighbour algorithm, implemented using the open-source TrackPy
package.\cite{allan_2021_4682814} We applied a constant-acceleration Kalman filter
with Rauch--Tung--Striebel (RTS) smoothing to the particle trajectories to mitigate
localization noise and obtain physically plausible velocity estimates for subsequent
model training (see Appendix~\ref{sec:kalman}).

\subsection*{Model Design}

Our framework integrates two coupled components: a Graph Network-based Simulator (GNS)
for particle velocity prediction, and a 3D U-Net for reconstructing multiphase fluid
interfaces. Both components are informed by pore geometry and exchange information at
each time step, enabling multimodal learning from experimental data.

\paragraph*{Graph Network-based Simulator (GNS):}
The GNS models particle dynamics represented in a Lagrangian
framework.\cite{10.5555/3524938.3525722} Each particle at time step $t$ is described
by a feature vector
\[
  \mathbf{x}_i^{(t)}=\left[ \mathbf{p}_i^{(t-1)}, \mathbf{p}_i^{(t)},
  \dot{\mathbf{p}}_i^{(t-1)}, \dots, \dot{\mathbf{p}}_i^{(t-C)} \right]
  \in \mathbb{R}^{3C+6},
\]
which contains the particle's positions at time steps $t-1$ and $t$, along with
historical velocity information from the past $C$ steps
$\{\dot{\mathbf{p}}_i^{(\tau)}\}_{\tau=t-C}^{t-1}$. Pairwise particle interactions
are encoded using a dynamically constructed graph, where edges are defined via a
radius-based connectivity criterion:
\[
  \mathcal{E}^{(t)} = \left\{ (i, j) \mid \lVert \mathbf{p}_i^{(t)} - \mathbf{p}_j^{(t)} \rVert_2 < r \right\},
\]
with $\mathbf{p}_i^{(t)} \in \mathbb{R}^3$ denoting the position of particle $i$, and
$r = 32$ voxels representing the interaction radius, selected based on physical flow
correlation length (see Fig.~\ref{fig:patchsize}A). For each edge, the raw features
$\mathbf{d}_{ij}^{(t)}$ include historical relative positional displacements and their
combined magnitude. Specifically,
\[
  \mathbf{d}_{ij}^{(t)} = \left[ \boldsymbol{\delta}_{ij}^{(t)},\ \lVert \boldsymbol{\delta}_{ij}^{(t)} \rVert_2 \right] \in \mathbb{R}^{7},
\]
where the historical relative displacement is
\[
  \boldsymbol{\delta}_{ij}^{(t)} = [(\mathbf{p}_i^{(t)} - \mathbf{p}_j^{(t)}),\ (\mathbf{p}_i^{(t-1)} - \mathbf{p}_j^{(t-1)})] \in \mathbb{R}^6.
\]

Notably, both raw node and edge features are normalized to have zero mean and unit
variance. After extracting raw features, the node and edge feature embeddings are
obtained via encoders $\phi_\text{node}$ and $\phi_\text{edge}$, respectively,
\textit{i.e.}
\[
  \mathbf{h}_i^{(t)} = \phi_\text{node}(\mathbf{x}_i^{(t)}), \qquad
  \mathbf{e}_{ij}^{(t)} = \phi_\text{edge}(\mathbf{d}_{ij}^{(t)}).
\]

Pore geometry information is incorporated via a convolutional neural network-based
image encoder $\phi_\text{image}(\mathcal{G}, \hat{\mathcal{S}}^{(t)})$, which
processes localized voxel patches extracted from the full CT scan of the dry pore
space $\mathcal{G}$, and current multiphase interface $\hat{\mathcal{S}}^{(t)}$
predicted by U-Net (see \textit{3D U-Net for Interface Prediction}). These patches,
centered on each particle, capture the nearby pore geometric environment relevant to
local flow behavior. The encoded geometry features are then added to particle
representations, enabling pore geometry-conditioned dynamics:
\[
  \mathbf{h}_i^{(t,0)} \leftarrow \mathbf{h}_i^{(t)} + \phi_\text{image}(\mathcal{G}, \hat{\mathcal{S}}^{(t)}),
\]
where $\mathbf{h}_i^{(t,0)}$ denotes the initial node representation before message
passing. The GNS backbone consists of a stack of $L=10$ message-passing layers.
Within each layer $l\in[0,9]$ at time $t$, edges features are first updated as
\[
  \mathbf{e}_{ij}^{(t,l)} \leftarrow \mathbf{e}_{ij}^{(t,l)} + \phi_\text{edge-update}(\mathbf{h}_i^{(t,l)}, \mathbf{h}_j^{(t,l)}, \mathbf{e}_{ij}^{(t,l)}),
\]
followed by the node update
\[
  \mathbf{h}_i^{(t,l)} \leftarrow \mathbf{h}_i^{(t,l)} + \phi_\text{node-update}\left(\mathbf{h}_i^{(t,l)}, \sum_{j \in \mathcal{N}(i)} \mathbf{e}_{ij}^{(t,l)} \right),
\]
where $\mathcal{N}(i)$ denotes the neighbors of node $i$. The outputs
$\mathbf{e}_{ij}^{(t,l)}$ and $\mathbf{h}_i^{(t,l)}$ serve as the input to the next
layer $l+1$. After $L$ iterations, the final node embeddings $\mathbf{h}_i^{(t,L)}$
are decoded to predict the current particle velocities:
\[
  \hat{\mathbf{v}}_i^{(t)} = \phi_\text{decode}(\mathbf{h}_i^{(t,L)}).
\]

Note that $\phi_\text{node}, \phi_\text{edge}, \phi_\text{edge-update},
\phi_\text{node-update}$ are all multi-layer perceptrons (MLPs). The model is trained
using a normalized mean squared error (MSE) loss between predicted and ground truth
particle velocities based on one-step prediction, while Gaussian noise is added to
particle positions to improve rollout robustness (see Appendix~\ref{sec:training} for
details). During inference, multi-step trajectories are generated autoregressively by
updating particle positions using a central-difference update. At each step, the graph
is reconstructed based on the updated positions, ensuring that message passing reflects
the evolving flow topology. This dynamic recomputation helps maintain physical
plausibility and prevents error accumulation in long rollouts.

\paragraph*{3D U-Net for Interface Prediction:}
The multiphase interface is modeled as a voxel occupancy field $\mathcal{S}^{(t)}$,
representing the non-wetting fluid phase at time $t$. The U-Net takes as input a
sequence of $N_\text{in}$ previous occupancy fields, along with the downsampled
voxelized velocity field interpolated from the GNS outputs. Input channels per voxel
include:
\[
  \mathcal{I} = \left\{ \mathcal{S}^{(t - N_\text{in} + 1)}, \ldots, \mathcal{S}^{(t)},
  \mathcal{V}_x^{(t)}, \mathcal{V}_y^{(t)}, \mathcal{V}_z^{(t)}, \mathcal{G} \right\},
\]
where $\mathcal{V}_x^{(t)}, \mathcal{V}_y^{(t)}, \mathcal{V}_z^{(t)}$ denote the
velocity components along the three orthogonal spatial directions and $\mathcal{G}$
is the static pore geometry mask, the model is trained using a voxel-wise binary
cross-entropy loss.

The network architecture follows an encoder-decoder design with skip connections. Each
downsampling stage consists of a 3D max-pooling layer followed by two 3D convolutions
with batch normalization and ReLU activations. The number of feature channels doubles
at each level of the encoder. The decoder mirrors this pattern using transposed
convolutions for upsampling and concatenating the corresponding encoder features to
retain spatial information. Formally, the prediction for the next interface frame is
given by:
\[
  \hat{\mathcal{S}}^{(t+1)} = \text{U-Net}(\mathcal{I}).
\]

To integrate particle-derived velocity information into the grid-based U-Net while
reducing memory cost, we apply an information-preserving pooling operation
(Fig.~\ref{fig:patchsize}B, see Appendix~\ref{sec:maxpool} for a detailed
description). This downsampling strategy, in contrast to na\"ive slicing or averaging,
selectively retains local velocity extremes within each pooling block, thereby
preserving more than 99\% of the particle information despite an eight-fold reduction
in resolution.

\subsection*{Multimodal Inference: Coupled Velocity and Interface Prediction}

The proposed framework combines particle-based velocity modeling with voxelized
interface prediction in a tightly coupled, multimodal setting. This allows for
coherent long-horizon rollouts of multiphase flow that respect both fluid dynamics and
evolving surface morphology.

At each time step $t$, the system predicts:
\begin{enumerate}
  \item Particle velocities $\hat{\mathbf{v}}_i^{(t)}$ using the Graph Network-based Simulator (GNS).
  \item The updated multiphase interface $\hat{\mathcal{S}}^{(t+1)}$ using a 3D U-Net conditioned on velocity fields and pore geometry.
\end{enumerate}

\begin{algorithm}[!tb]
  \caption{Multimodal Rollout of Velocity and Interface Evolution}
  \begin{algorithmic}[1]
    \State \textbf{Input:} Initial particle position history $\{\mathbf{p}_i^{(t)}\}_{t=\tau-1}^{\tau}$, velocity history $\{\dot{\mathbf{p}}_i^{(t)}\}_{t=\tau-C}^{\tau-1}$, interface history $\{\mathcal{S}^{(t)}\}_{t=\tau - N_\mathrm{in} + 1}^{\tau}$, pore geometry $\mathcal{G}$, and graph radius $r$

    \For{each time step $t = \tau, \tau+1, \tau+2, \ldots$}

    \State \textbf{Predict particle velocities} using GNS:
    \[
      \hat{\mathbf{v}}_i^{(t)} \leftarrow \mathrm{GNS}\left( \{\mathbf{p}_i^{(t)}\}, \{\dot{\mathbf{p}}_i^{(t)}\},  \{\mathcal{S}^{(t)}\}, \mathcal{G}\right)
    \]

    \State \textbf{Update particle positions}:
    \[
      \mathbf{p}_i^{(t+1)} \leftarrow \mathbf{p}_i^{(t-1)} + 2 \, \hat{\mathbf{v}}_i^{(t)}
    \]

    \State \textbf{Recompute graph structure}:
    \[
      \mathcal{E}^{(t+1)} \leftarrow \left\{ (i, j) \, \middle| \, \| \mathbf{p}_i^{(t+1)} - \mathbf{p}_j^{(t+1)} \|_2 < r \right\}
    \]

    \State \textbf{Interpolate velocities to voxel grid}:
    \[
      \mathcal{V}^{(t)} \leftarrow \text{Interpolate}\left( \mathbf{p}_i^{(t+1)}, \hat{\mathbf{v}}_i^{(t)} \right)
    \]

    \State \textbf{Prepare multimodal input for U-Net}:
    \[
      \mathcal{I} \leftarrow \left\{ \mathcal{S}^{(t - N_\mathrm{in} + 1)}, \ldots, \mathcal{S}^{(t)}, \, \mathcal{V}_x^{(t)}, \mathcal{V}_y^{(t)}, \mathcal{V}_z^{(t)}, \, \mathcal{G} \right\}
    \]

    \State \textbf{Predict next interface using U-Net}:
    \[
      \hat{\mathcal{S}}^{(t+1)} \leftarrow \mathrm{U\text{-}Net}(\mathcal{I})
    \]

    \State \textbf{Update interface history with} $\hat{\mathcal{S}}^{(t+1)}$

    \State \textbf{Condition next GNS input on} $\hat{\mathcal{S}}^{(t+1)}$

    \EndFor
  \end{algorithmic}
  \label{alg:multimodal_rollout}
\end{algorithm}

Algorithm~\ref{alg:multimodal_rollout} outlines the workflow. The coupled architecture
respects both Lagrangian particle motion and Eulerian voxel-based interface
representations.

\subsection*{Model Ablation and Variants}

To assess the robustness of our architecture, we performed the following ablation
experiments.

\begin{figure}[!tb]
  \centering
  \includegraphics[width=\columnwidth]{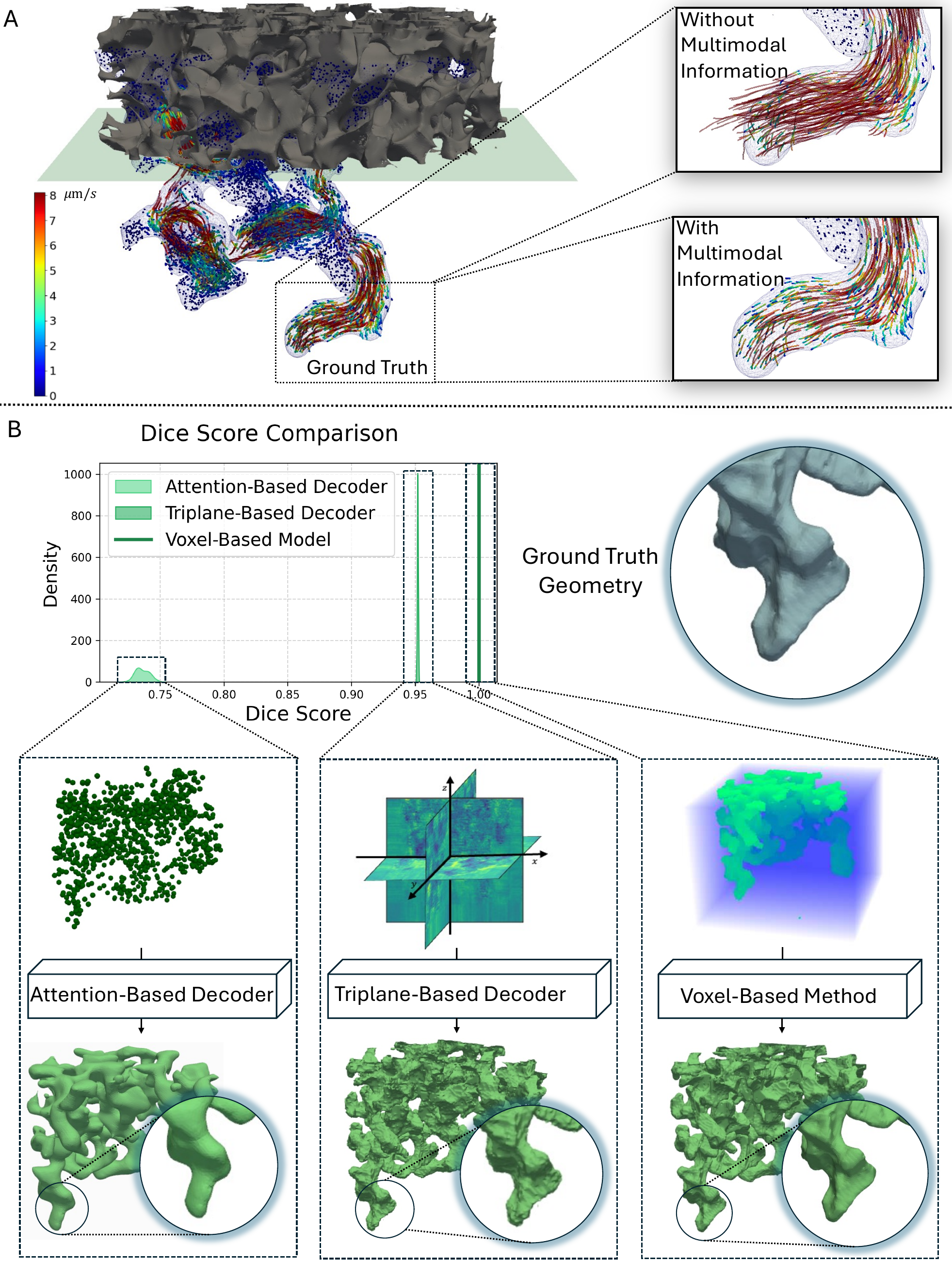}
  \caption{\textbf{Ablation Experiments.}
    \textbf{(A)} The sintered glass matrix (gray) under the green plane is made
    transparent to expose the non-wetting phase (blue). Velocity tracer trajectories
    are color-mapped by their instantaneous velocity magnitude. When interface and pore
    structure inputs are removed from the GNS, particles exhibit unphysical behavior,
    including escaping beyond the fluid interface in the Haines jump region.
    \textbf{(B)} Geometry representation benchmarks: triplane and VecSet degrade
    interface detail, while voxel-based U-Net reconstructions preserve ground truth
    geometry.}
  \label{fig:geo_cmp}
\end{figure}

We first examined the performance of the GNS model when it is stripped of information
about the multiphase interface and the solid pore structure. In the full multimodal
formulation, the model takes as input particle states together with pore geometry and
interface information, which we write as
$\mathrm{GNS}(\{\mathbf{p}_i^{(t)}\}, \{\mathcal{S}^{(t)}\}, \mathcal{G})$; in the
ablated setting, the model is reduced to $\mathrm{GNS}(\{\mathbf{p}_i^{(t)}\})$,
receiving no explicit geometric conditioning. This simplification produced strongly
unphysical behavior: particle trajectories were predicted to escape from the nonwetting
phase region, a violation of the governing physics (Fig.~\ref{fig:geo_cmp}A). This
experiment demonstrates that coupling GNS to both solid structure and dynamic interface
geometry is indispensable for consistent rollout predictions.

Multiphase interface geometry plays a critical role in predicting the evolution of
particle trajectories. In Fig.~\ref{fig:geo_cmp}B, we investigated how best to
represent this information. To this end, we benchmarked the U-Net model
architecture\cite{Ronneberger_Fischer_Brox_2015} against several of the most recent
advanced approaches.

We evaluated two classes of learned 3D representations: triplane-based encoder-decoder
and VecSet-based surface representations. In triplane models, a 3D point
$\mathbf{x} \in \mathbb{R}^3$ is projected onto three orthogonal 2D planes
$(xy, xz, yz)$, and its occupancy is predicted as
$\hat{\mathbf{O}} = NN_\phi(\mathbf{x}, \mathbf{f}_{xy}, \mathbf{f}_{xz}, \mathbf{f}_{yz})$,
where $\{\mathbf{f}_{xy}, \mathbf{f}_{xz}, \mathbf{f}_{yz}\}$ are learned feature
maps encoding geometry from each plane.\cite{Peng_Niemeyer_Mescheder_Pollefeys_Geiger_2020, Shue_Chan_Po_Ankner_Wu_Wetzstein_2022}
VecSet models, in contrast, encode the geometry as a set of latent vectors
$\{\mathbf{z}_1, ..., \mathbf{z}_N\}$ derived from sampled surface
points.\cite{Zhang_Tang_Niessner_Wonka_2023} The occupancy of a query point is
decoded via attention-based aggregation over these latent codes. Both models were
trained on 150 samples and tested on 50 unseen geometries. Qualitatively, triplane
reconstructions were noisy and lacked local consistency, while VecSet results were
overly smoothed and failed to resolve detailed curvature. Quantitatively, Dice
similarity scores averaged 0.952 for triplane and 0.736 for VecSet, while the U-Net
model consistently achieved scores approaching unity.

%%% FIGURE 5 %%%

To investigate how the velocity information from GNS model affects multiphase interface
prediction, the full model---which incorporates interface history, velocity field, and
pore geometry:
\[
  \text{U-Net}\left(\left\{ \mathcal{S}^{(t - N_\text{in} + 1)}, \ldots, \mathcal{S}^{(t)}, \mathcal{V}_x^{(t)}, \mathcal{V}_y^{(t)}, \mathcal{V}_z^{(t)}, \mathcal{G} \right\}\right),
\]
is compared against a na\"ive variant that omits the velocity field:
\[
  \text{U-Net}\left(\left\{ \mathcal{S}^{(t - N_\text{in} + 1)}, \ldots, \mathcal{S}^{(t)}, \mathcal{G} \right\}\right).
\]

The na\"ive model generated non-physical artifacts such as disconnected fluid regions
and inconsistent boundary evolution. Including the velocity field improved the Dice
score from 0.84 to 0.94 and eliminated these pathologies, highlighting the importance
of velocity in interface prediction, as shown in Fig.~\ref{fig:result}C.

These ablations highlight the interdependence between accurate geometric conditioning
and reliable interface representation. The performance of GNS degrades significantly
without explicit access to interface and structure information, and our benchmarks show
that among candidate representations, voxel-based U-Nets best preserve the geometric
fidelity required for physically plausible predictions. This tight coupling between
accurate geometry and consistent rollout dynamics is especially critical in porous
media, where small-scale interface features directly influence capillary phenomena and
phase behavior.\cite{Armstrong_McClure_Robins_Liu_Arns_Schlüter_Berg_2019}

\section*{Author Contributions}

C.W., L.Z., Y.G., and G.W. designed research; C.W., Y.G., L.Z., and G.W. performed
research; R.v.d.M. and T.B. contributed data and interpretation; C.W., Y.G., L.Z.,
X.J., C.S., and R.Y. contributed to code development; C.W. and Y.G. analyzed data;
G.W., T.P., S.K., and M.J.B. supervised the research; G.W. project administration,
funding acquisition; C.W. wrote the original draft. All authors reviewed and edited
the draft.

\section*{Conflicts of Interest}

The authors declare no competing interest.

\section*{Data Availability}

Source code for the proposed model, training and benchmarking code are available at
GitHub (\url{https://github.com/chunyang-w/pore_gns}). The pore-scale experimental
datasets used in this study are publicly available in the PSI Public Data Repository
(\url{https://doi.org/10.16907/c0dfa6c8-25da-454e-82fa-fc5db7f7c6f2}).

\section*{Acknowledgements}

C.W.\ acknowledges the support of the Department of Earth Science and Engineering,
Imperial College London, through a Janet Watson PhD Fellowship. L.Z.\ and G.W.\
acknowledge the generous support of Schmidt Sciences through the AI2050 fellowship.
T.B.\ and R.v.d.M.\ are supported by an ERC grant (FLOWSCOPY, 101116228) funded by
the European Union. Views and opinions expressed are, however, those of the author(s)
only and do not necessarily reflect those of the European Union or the European
Research Council Executive Agency. Neither the European Union nor the granting
authority can be held responsible for them. The experimental data reused in this work
were acquired at the TOMCAT beamline X02DA of the SLS, Paul Scherrer Institut,
Villigen, Switzerland (beam time proposal 20212066).

%%%=============================================================
%%% REFERENCES
%%%=============================================================

\bibliography{references}
\bibliographystyle{unsrtnat}

%%%=============================================================
%%% ELECTRONIC SUPPLEMENTARY INFORMATION (Appendix)
%%%=============================================================

\onecolumn  % Switch to single column for ESI

\vspace{2em}
\hrule
\vspace{1em}

\begin{center}
  {\LARGE\textbf{Appendix}}\\[0.5em]
  {\large Learning Pore-scale Multiphase Flow from 4D Velocimetry}
\end{center}

\vspace{1em}
\hrule
\vspace{2em}

% Reset counters with S prefix for ESI
\renewcommand{\thefigure}{S\arabic{figure}}
\renewcommand{\thetable}{S\arabic{table}}
\renewcommand{\thesection}{S\arabic{section}}
\renewcommand{\theequation}{S\arabic{equation}}
\setcounter{figure}{0}
\setcounter{table}{0}
\setcounter{section}{0}
\setcounter{equation}{0}

\section{Training Details and Model Parameters}
\label{sec:training}

The particle-scale velocity prediction component of our framework is based on a Graph
Neural Network (GNS) architecture trained using an autoregressive rollout strategy.
Node features consist of two-step particle positions (6 features) and a velocity
history over $C=5$ steps (15 features), resulting in a total of 21 input features per
node. Edge attributes include two-step pairwise relative positions (6 features) and
their $\ell_2$ norm (1 feature), yielding 7 edge features per edge. A graph is
constructed at each time step using a radius-based criterion with $r = 32$ voxels and
a maximum of 64 neighbors per node.

The GNS employs 10 message-passing layers, each comprising residual connections and
fully connected update functions with 128 hidden dimensions. A CNN-based encoder
integrates local geometry information extracted from voxel patches centered on each
particle, downsampled from the original CT data. The final decoder predicts
three-dimensional particle velocities at the next time step.

Data normalization is applied based on statistics computed from the entire training
set. Models are optimized with Adam optimizer, with a learning rate of
$5 \times 10^{-5}$, weight decay of $5 \times 10^{-4}$, and a cosine annealing
learning rate schedule with $T_{\mathrm{max}} = 200$ epochs. Models are trained for
600 epochs with a batch size of 1, on a single NVIDIA A100 (80GB) GPU.

To improve rollout robustness, we adopt a noise injection strategy similar to that
used in the original GNS framework. Specifically, during training, Gaussian noise is
added to the ground truth particle positions used to construct the input graph,
encouraging the model to learn stable dynamics under small perturbations. This helps
mitigate compounding errors during autoregressive inference.

The interface morphology component uses a 3D U-Net trained to predict binary fluid
occupancy at the next time step, conditioned on a history of $N_{\mathrm{in}} = 2$
previous segmented interface fields, corresponding velocity fields, and the static
pore geometry. Input channels include three velocity components, one binary interface
channel per frame, and an optional pore geometry mask. The network employs a standard
encoder-decoder structure with skip connections, trained using binary cross-entropy
loss.

U-Net is trained for 100 epochs with a batch size of 2, using the Adam optimizer with
a learning rate of $10^{-3}$. Mixed-precision training with automatic casting to FP16
is enabled to improve computational efficiency. Data downsampling by a factor of 8 is
applied to both velocity and geometry fields for memory efficiency. The dataset is
divided temporally, with 75\% used for training and 25\% reserved for testing.

\section{Failure Mode Analysis}
\label{sec:failure_modes}

\begin{figure}[!tbp]
  \centering
  \includegraphics[width=\textwidth]{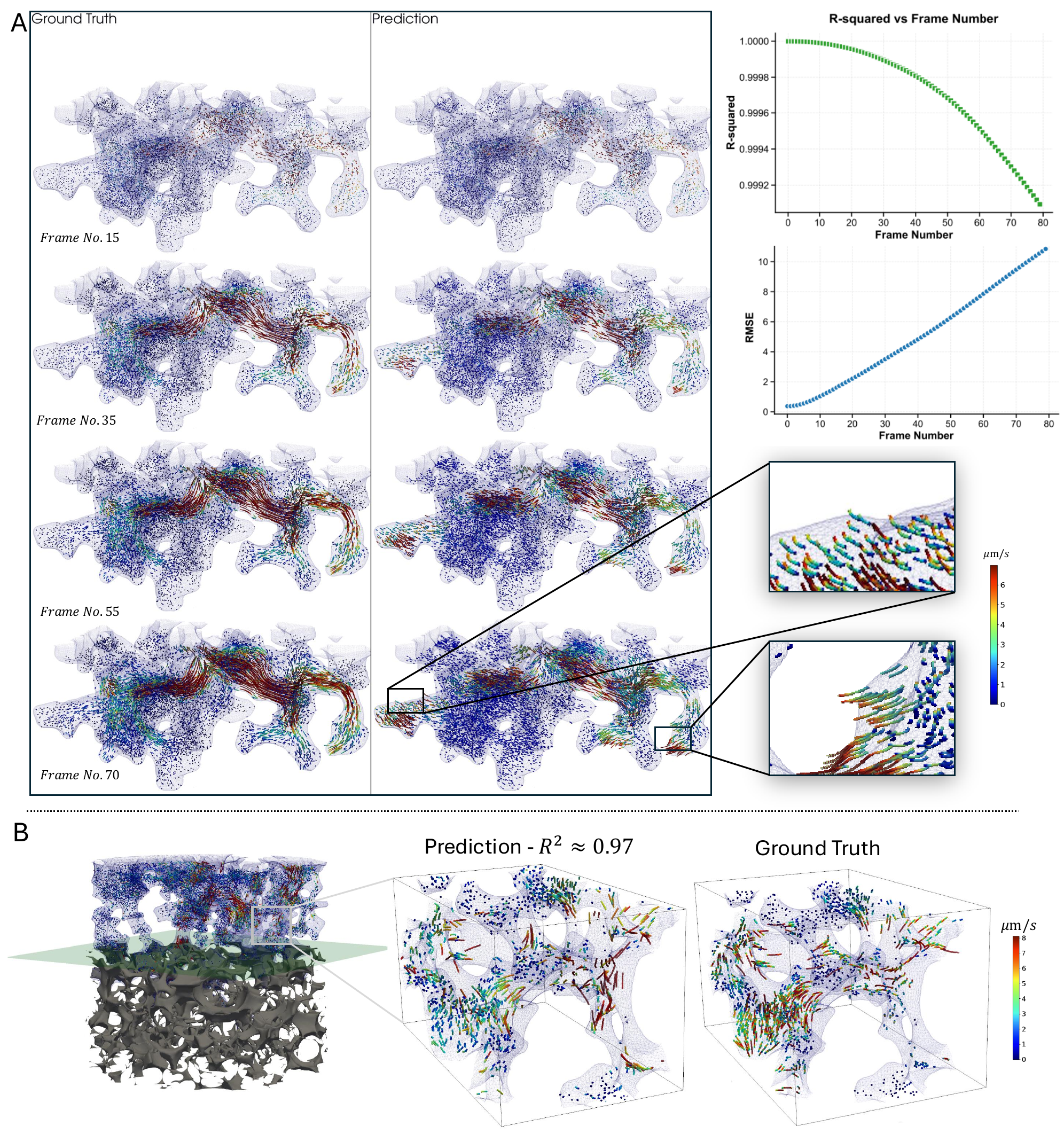}
  \caption{\textbf{Failure modes of the learned rollout under long-horizon prediction
    and cross-rock generalization.}
    \textbf{(A)} Extended rollout failure. Predicted interface evolution and particle
    velocities remain consistent with the ground truth at early timesteps
    (frames 15--45), but visible deviations emerge by frame 70 as errors accumulate
    autoregressively. This degradation is reflected in the increasing RMSE and
    decreasing $R^2$ over time, particularly toward the end of the rollout.
    \textbf{(B)} Generalization-induced failure under cross-rock shift. Zero-shot
    inference on a Ketton limestone sample using a model trained exclusively on
    sintered-glass data. While the overall velocity field and particle statistics
    remain well correlated with the ground truth, localized regions exhibit degraded
    trajectory tracking and velocity alignment, highlighting limitations under
    pronounced pore-structure domain shift.}
  \label{fig:failure}
\end{figure}

Fig.~\ref{fig:failure} illustrates two representative failure modes of the learned
framework under extended autoregressive rollout and cross-rock domain shift. During
long-horizon prediction (Fig.~\ref{fig:failure}A), the model maintains strong
agreement with the ground truth at early timesteps (frames 15--45), but small velocity
and interface discrepancies accumulate over time, leading to visible deviations by
frame 70. This temporal degradation is reflected in the gradual increase in RMSE and
decrease in $R^2$ toward the end of the rollout.

Under strict zero-shot inference on the Ketton limestone sample
(Fig.~\ref{fig:failure}B), the global velocity statistics remain well correlated with
experimental measurements; however, localized regions exhibit reduced trajectory
alignment and flow direction consistency. These failure modes highlight the sensitivity
of autoregressive dynamics to compounding errors and the challenges posed by pronounced
pore-structure domain shifts.

\section{Supplementary Training Data}
\label{sec:supp_training}

To improve in-domain generalization and assess robustness to experimental variability,
we augment the training set with two additional drainage experiments acquired from the
same sintered-glass sample used in Exp.~$\alpha$. These experiments, denoted
Exp.~$\alpha'$ and Exp.~$\alpha''$, are performed on the identical porous medium,
ensuring that all three datasets share the same pore geometry.

Both supplementary experiments are conducted at an injection rate of 348~nL/min.
Exp.~$\alpha'$ is acquired at a temporal resolution of 0.5~s per frame, yielding 100
frames. To maintain consistency with the main dataset, the sequence is temporally
interpolated to a resolution of 0.25~s per frame. Exp.~$\alpha''$ is recorded directly
at 0.25~s per frame with 100 frames captured.

\begin{figure}[htbp]
  \centering
  \includegraphics[width=\textwidth]{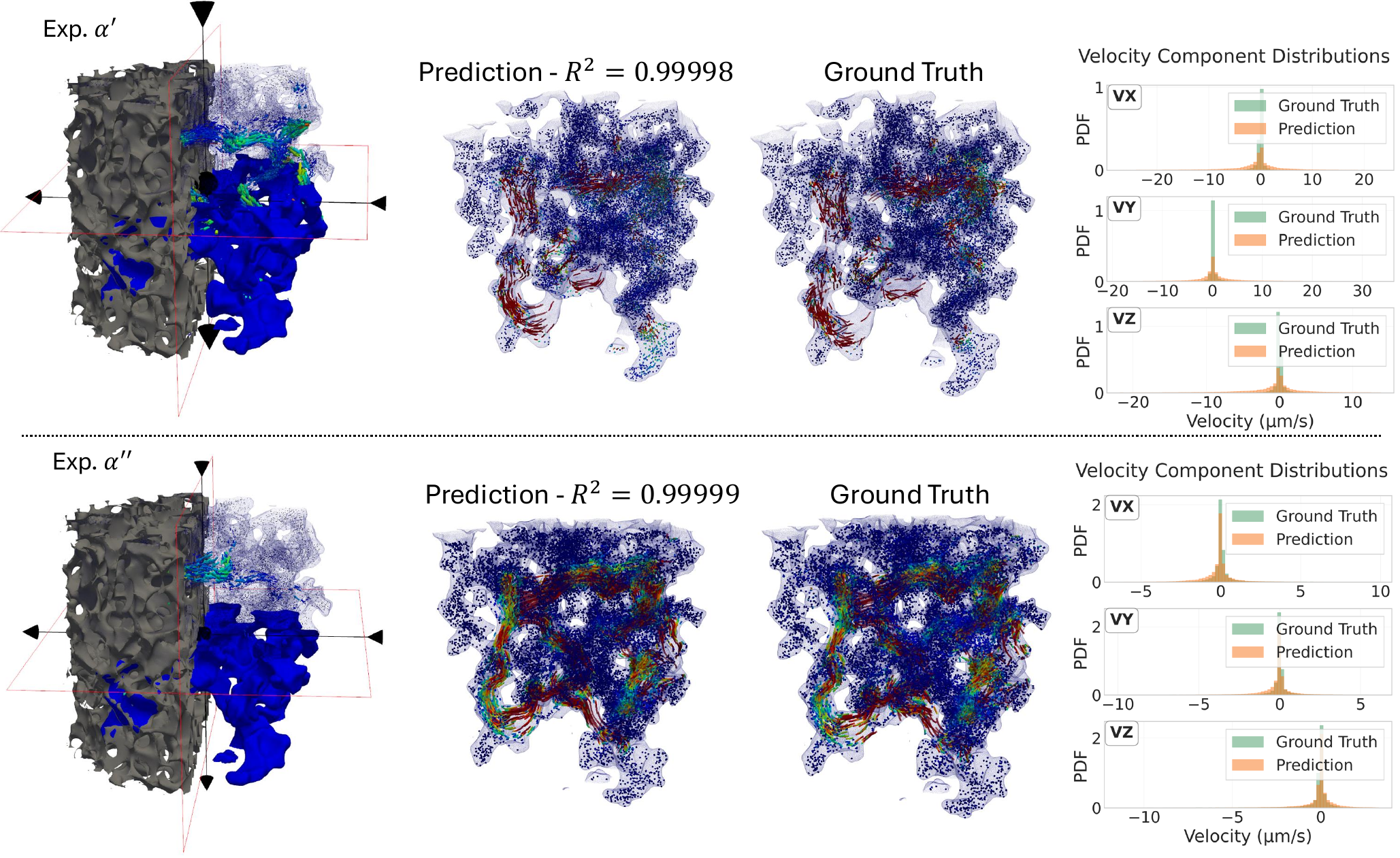}
  \caption{Supplementary training data for improved in-domain generalization. Two
    additional drainage experiments from the sintered-glass sample used in
    Exp.~$\alpha$ (Exp.~$\alpha'$ and Exp.~$\alpha''$) are incorporated into training.
    The data are split into two folds, with performance evaluated on the corresponding
    held-out test sets. Reported metrics include $R^{2}$ computed from tracer
    trajectory agreement and comparisons of velocity-component distributions between
    predictions and ground truth for both experiments.}
  \label{fig:extra_exp}
\end{figure}

The model is trained jointly on Exp.~$\alpha$, Exp.~$\alpha'$, and Exp.~$\alpha''$.
For each dataset, the final 25\% of frames in time are held out and used as test data,
while the remaining frames are used for training. Performance on the held-out portions
of Exp.~$\alpha'$ and Exp.~$\alpha''$ is reported in Fig.~\ref{fig:extra_exp},
including trajectory-based $R^{2}$ metrics and comparisons of velocity-component
distributions.

The model trained on the combined sintered-glass dataset is subsequently used, without
any fine-tuning, to perform strict zero-shot inference on the Ketton limestone
experiment described in the main text.

\section{Max-Pooling for Velocity Field Downsampling}
\label{sec:maxpool}

Efficient downsampling of particle velocity data is essential for integrating
Lagrangian particle information into grid-based learning frameworks while controlling
memory usage. In our workflow, particle velocities are first interpolated onto a
regular voxel grid that spans the pore space. To further reduce resolution without
excessive information loss, we apply a custom three-dimensional pooling operation.

The pooling is performed over cubic blocks of size $f \times f \times f$, where $f$
is the downsampling factor. For each block, the output value is determined by
selecting the element (maximum or minimum) with the larger absolute value:
\[
  \mathcal{P}_{ijk} =
  \begin{cases}
    \max(\mathcal{B}_{ijk}) & \text{if} \ \left| \max(\mathcal{B}_{ijk}) \right| \geq \left| \min(\mathcal{B}_{ijk}) \right|, \\
    \min(\mathcal{B}_{ijk}) & \text{otherwise}.
  \end{cases}
\]

Here, $\mathcal{B}_{ijk}$ represents the set of velocity values within the
$(i,j,k)$-th pooling block, and $\mathcal{P}_{ijk}$ is the pooled result. This
approach ensures that dominant features with high magnitude, such as regions of strong
flow, sharp gradients, or boundary layer structures, are preferentially retained during
the downsampling process.

Na\"ive downsampling techniques, such as uniform subsampling or simple averaging, are
known to cause severe information loss in three-dimensional datasets, particularly when
representing sparse or high-contrast physical fields as in particle velocity. Such
methods often fail to preserve critical flow features, resulting in degraded model
inputs and reduced prediction accuracy. The pooling operation adopted here mitigates
this problem by selectively preserving velocity extremes within each spatial block,
maintaining essential flow information while reducing data dimensionality.

In practice, we apply this pooling operation with a downsampling factor of $f = 8$,
which provides a favorable balance between memory efficiency and preservation of key
velocity structures critical for learning accurate flow-interface interactions.

\section{Flow Field Reconstruction Details}
\label{sec:flow_reconstruction}

To evaluate model-predicted Lagrangian particle velocities against voxel-based
experimental flow fields, we reconstruct a dense Eulerian velocity field via
interpolation. Given a set of predicted particle velocities $\{\mathbf{v}_i\}$ at
spatial positions $\{\mathbf{x}_i\}$ within the pore space, we interpolate onto a
uniform voxel grid using a physically constrained method.

For each velocity component, we augment the interpolation set with boundary points
sampled from the fluid--solid interface, identified by morphological dilation of the
solid-phase mask. These boundary points $\{\mathbf{x}_b\}$ are assigned zero velocity
to enforce the no-slip condition. The interpolation problem then becomes:
\[
  \mathbf{v}(\mathbf{x}) = \mathrm{Interp}(\{\mathbf{x}_i, \mathbf{x}_b\}, \{\mathbf{v}_i, \mathbf{0}\}; \mathbf{x}),
\]
where $\mathrm{Interp}$ denotes trilinear interpolation over the combined particle and
boundary constraints.

A tunable ratio $\rho$ controls the number of boundary points relative to
velocity-carrying particles, balancing computational efficiency and fidelity near
walls. The resulting velocity field is masked by the pore geometry to remove any
spurious values in the solid domain. This procedure yields a continuous voxel-based
velocity field suitable for direct comparison with experimental measurements using
standard metrics such as MAE and spatial correlation.

\section{SDF-Based Interpolation of Fluid Interfaces Between Temporal Frames}
\label{sec:sdf_interp}

To bridge the temporal resolution mismatch between training (0.25 s) and test data
(0.5 s), we interpolate intermediate fluid interfaces between observed frames. This
allows autoregressive rollout at the model's native time scale using temporally
consistent inputs.

Given binary pore segmentation $V_0$ and $V_1$ at consecutive time steps, we compute
their corresponding signed distance fields (SDFs), defined as
\[
  \phi_i(\mathbf{x}) = d_{\text{in}}^i(\mathbf{x}) - d_{\text{out}}^i(\mathbf{x}), \quad i \in \{0, 1\},
\]
where $d_{\text{in}}^i$ and $d_{\text{out}}^i$ are Euclidean distance transforms of
the foreground and background regions, respectively. Intermediate geometries are
obtained by linear interpolation in SDF space:
\[
  \phi_t(\mathbf{x}) = (1 - t)\phi_0(\mathbf{x}) + t\phi_1(\mathbf{x}), \quad t \in [0, 1],
\]
followed by thresholding $\phi_t(\mathbf{x}) \geq 0$ to produce a binary mask. This
method ensures smooth geometric transitions between frames and preserves physical
continuity in interface evolution.

\section{Trajectory Smoothing Using a Constant-Acceleration Kalman Smoother}
\label{sec:kalman}

Raw particle trajectories obtained from micro-velocimetry contain localization noise
due to imaging artifacts, sub-voxel interpolation, and tracking uncertainty.
Differentiating such noisy positions to obtain velocities amplifies high-frequency
errors, producing non-physical velocity spikes and jitter that can dominate the
learning signal in data-driven models.

To obtain smoother trajectories while preserving transient dynamics, we employ a
model-based state-space approach: a 3D constant-acceleration (CA) Kalman filter
followed by Rauch--Tung--Striebel (RTS) backward smoothing. The CA model treats each
particle as evolving according to
\[
  \mathbf{s}_t = [x_t, \dot{x}_t, \ddot{x}_t,\ y_t, \dot{y}_t, \ddot{y}_t,\ z_t, \dot{z}_t, \ddot{z}_t]^\top \in \mathbb{R}^9,
\]
with a discrete-time transition
\[
  \mathbf{s}_{t+1} = \mathbf{F}(\Delta t)\,\mathbf{s}_t + \mathbf{w}_t,
\]
where $\mathbf{F}(\Delta t)$ is the standard CA kinematic matrix (position integrates
velocity and acceleration, and velocity integrates acceleration), and
$\mathbf{w}_t \sim \mathcal{N}(\mathbf{0},\mathbf{Q})$ represents process noise. In
our implementation, $\mathbf{Q}$ is derived from a white-jerk assumption (acceleration
driven by random jerk), which permits smooth but flexible changes in acceleration
between frames.

The measurement model uses only observed positions,
\[
  \mathbf{z}_t = \mathbf{H}\,\mathbf{s}_t + \mathbf{v}_t, \qquad
  \mathbf{H}\mathbf{s}_t = [x_t, y_t, z_t]^\top,
\]
with $\mathbf{v}_t \sim \mathcal{N}(\mathbf{0},\mathbf{R})$ representing measurement
noise. The forward Kalman recursion produces filtered state estimates and covariances
that optimally fuse the motion prior with noisy observations under Gaussian
assumptions. RTS smoothing then performs a backward pass to compute a globally
consistent trajectory estimate that leverages information from both past and future
frames, substantially reducing lag and edge artifacts relative to causal filters.

This Kalman smoother is well matched to our application because it (i)~enforces a
physically interpretable kinematic prior (continuous velocity and acceleration); and
(ii)~directly outputs denoised positions and velocities in a single procedure.

\section{Velocity Prediction Error Metric}
\label{sec:nrmse_metric}

We report velocity prediction accuracy using $\mathrm{NRMSE_{p99}}$, defined below,
to provide a stable reference scale for comparison across datasets. Standard
per-particle mean relative error (MRE) can be sensitive to capillary-dominated
drainage experiments, where a substantial fraction of tracer particles are effectively
stagnant between drainage events. In the Ketton limestone dataset, for example, 43.9\%
of particle-frame observations have a ground-truth velocity magnitude below
0.1~vox/frame, and the corresponding fraction is 48.2\% for the in-distribution
sintered-glass dataset (Exp.~$\alpha$). Both velocity distributions are strongly
right-skewed: mean-to-median ratios are $4.7$ for the sintered-glass dataset and $1.9$
for Ketton, confirming that the typical particle is effectively stagnant even as a
small fraction of fast-moving particles drives the drainage dynamics. Velocities at or
below 0.1~vox/frame fall within the experimental precision, so relative-error
calculations in this regime can be dominated by measurement noise and are not preferred
for error assessment. For such particles, the denominator of the relative error,
\[
  \epsilon_{\mathrm{rel},i} = \frac{|\,\|\hat{\mathbf{v}}_i\| - \|\mathbf{v}_{\mathrm{gt},i}\|\,|}{\|\mathbf{v}_{\mathrm{gt},i}\|},
\]
approaches zero, which can produce large $\epsilon_{\mathrm{rel},i}$ values even when
the absolute prediction error is sub-voxel. These near-stagnant particles are less
informative for the model's capacity to predict advective transport: their positional
uncertainty is dominated by sub-voxel localisation noise and tracking artefacts rather
than by dynamically meaningful flow. Accordingly, we prioritize metrics that reflect
advective transport rather than measurement noise in near-stagnant regions.

To obtain a physically meaningful and numerically stable accuracy measure, we adopt
the normalised root-mean-square error relative to the 99th-percentile characteristic
velocity ($\mathrm{NRMSE_{p99}}$), defined as
\begin{equation}
  \mathrm{NRMSE_{p99}} = \frac{\mathrm{RMSE}(\|\mathbf{v}\|)}{Q_{99}(\|\mathbf{v}_{\mathrm{gt}}\|)},
  \qquad
  \mathrm{RMSE}(\|\mathbf{v}\|) = \sqrt{\frac{1}{N}\sum_{i=1}^{N}
    \bigl(\|\hat{\mathbf{v}}_i\| - \|\mathbf{v}_{\mathrm{gt},i}\|\bigr)^{2}},
  \label{eq:nrmse_p99}
\end{equation}
where $\hat{\mathbf{v}}_i$ and $\mathbf{v}_{\mathrm{gt},i}$ are the predicted and
ground-truth velocity vectors for particle $i$, $N$ is the total number of particles,
and $Q_{99}(\|\mathbf{v}_{\mathrm{gt}}\|)$ denotes the 99th percentile of the
ground-truth velocity-magnitude distribution computed over all particles and frames.
This metric has three desirable properties.

\begin{enumerate}
  \item \textbf{All particles are included.} Unlike threshold-based filters, which
    discard slow-moving particles and introduce selection bias,
    $\mathrm{NRMSE_{p99}}$ uses the full particle set. Near-stagnant particles
    contribute a small absolute error to the RMSE numerator proportional to their
    actual prediction error, rather than inflating a relative denominator.

  \item \textbf{Physically motivated reference scale.} The 99th-percentile velocity
    $Q_{99}$ is a reasonable estimate of the characteristic fast-particle velocity---
    the upper end of the advective flow regime---without being dominated by isolated
    outliers. Normalising by this quantity expresses the prediction error as a fraction
    of the characteristic flow speed, analogous to normalising by the free-stream
    velocity $U_\infty$ in external-flow computational fluid dynamics validation.

  \item \textbf{Robustness to distributional skewness.} Velocity magnitudes in
    porous-media flow are highly right-skewed: a few fast particles in preferential
    flow paths coexist with a majority of slow or stagnant particles. The 99th
    percentile captures the scale of the fast-flow regime without being inflated by
    extreme outliers (as the maximum would be) or suppressed by the stagnant majority
    (as the mean would be).
\end{enumerate}

For the Ketton limestone zero-shot dataset (frames 150--180, $N = 380\,370$
particles), the ground-truth 99th-percentile velocity magnitude is
$Q_{99} = 1.315$~vox/frame, and the RMSE is $0.2150$~vox/frame, yielding
$\mathrm{NRMSE_{p99}} = 16.4\%$. For reference, the in-distribution sintered-glass
dataset (Exp.~$\alpha$) has a 99th-percentile velocity of $Q_{99} = 4.700$~vox/frame,
reflecting the substantially higher characteristic flow speeds in that experiment.

\end{document}